%% file: CCA_TMM.tex
\documentclass[journal]{IEEEtran}
\ifCLASSINFOpdf
\else
\fi

\usepackage[colorlinks, citecolor=black]{hyperref}
\usepackage[numbers,sort&compress]{natbib}
\usepackage{amsmath,amsfonts}
\usepackage{algorithmic}
\usepackage{algorithm}
\usepackage{array}
\usepackage[caption=false,font=normalsize,labelfont=sf,textfont=sf]{subfig}
\usepackage{textcomp}
\usepackage{stfloats}
\usepackage{url}
\usepackage{verbatim}
\usepackage{graphicx}
\usepackage{verbatim}
\usepackage{multirow}
\usepackage{xcolor}
\usepackage{hyperref}
\newlength\secmargin
\newlength\subsecmargin
\newlength\paramargin
\newlength\figmargin
\newlength\eqmargin
\newcommand{\figref}[1]{Figure~\ref{fig:#1}}
\newcommand{\tabref}[1]{Table~\ref{tab:#1}}

\newcommand{\eg}{\textit{e}.\textit{g}.}
\newcommand{\ie}{\textit{i}.\textit{e}.}
\newcommand{\etal}{\textit{et al}. }
\newcommand{\Paragraph}[1]{\noindent\textbf{#1}}

\usepackage{color}
\begin{document}
		%
		\title{Learning Commonsense-aware Moment-Text Alignment for Fast Video Temporal Grounding}
		
		
		%
		%
		%
		
		\author{Ziyue~Wu,
			Junyu~Gao,
			Shucheng Huang, 
			and~Changsheng~Xu,~\IEEEmembership{Fellow,~IEEE}
			\IEEEcompsocitemizethanks{
				Ziyue Wu is with Tianjin University of Technology, Tianjin 300384, China, and also with the National Laboratory of Pattern Recognition, Institute of Automation, Chinese Academy of Sciences, Beijing 100190, China. (email: wuziyuewzy@gmail.com).

				Junyu Gao and Changsheng Xu are with the National Laboratory of Pattern Recognition, Institute of Automation, Chinese Academy of Sciences, Beijing 100190, China, and with School of Artificial Intelligence, University of Chinese Academy of Sciences, Beijing 100049, China. Changsheng Xu is also with the PengCheng Laboratory, Shenzhen 518066, China. (e-mail: junyu.gao@nlpr.ia.ac.cn; csxu@nlpr.ia.ac.cn).
						
				Shucheng Huang is with Jiangsu University of Science and Technology, Zhenjiang 212100. (email: schuang@just.edu.cn).}

			\thanks{Copyright (c) 2021 IEEE. Personal use of this material is permitted. However, permission to use this material for any other purposes must be obtained from the IEEE by sending a request to pubs-permissions@ieee.org.}
		}

\maketitle

\begin{abstract}
	\input{abst}
\end{abstract}


%
\IEEEpeerreviewmaketitle

\section{Introduction}
\label{sec:intro}
\input{intro}
\section{Related Work}
\label{sec:related}
\input{rw}

\section{Our approach}
\label{sec:method}
\input{ours}

\section{Experiments}
\label{sec:exp}
\input{expr}
\section{Conclusion}
\label{sec:conclusion}
\input{conclus}

\ifCLASSOPTIONcaptionsoff
\newpage
\fi



\small
\bibliographystyle{IEEEtran}
\bibliography{ieee2022bib}
%
%
%

%






\end{document}

%% file: abst.tex
Grounding temporal video segments described in natural language queries effectively and efficiently is a crucial capability needed in vision-and-language fields. In this paper, we deal with the fast video temporal grounding (FVTG) task, aiming at localizing the target segment with high speed and favorable accuracy. Most existing approaches adopt elaborately designed cross-modal interaction modules to improve the grounding performance, which suffer from the test-time bottleneck. Although several common space-based methods enjoy the high-speed merit during inference, they can hardly capture the comprehensive and explicit relations between visual and textual modalities. In this paper, to tackle the dilemma of speed-accuracy tradeoff, we propose a commonsense-aware cross-modal alignment (CCA) framework, which incorporates commonsense-guided visual and text representations into a complementary common space for fast video temporal grounding. Specifically, the commonsense concepts are explored and exploited by extracting the structural semantic information from a language corpus. Then, a commonsense-aware interaction module is designed to obtain bridged visual and text features by utilizing the learned commonsense concepts. Finally, to maintain the original semantic information of textual queries, a cross-modal complementary common space is optimized to obtain matching scores for performing FVTG. Extensive results on two challenging benchmarks show that our CCA method performs favorably against state-of-the-arts while running at high speed. Our code is available at \href{https://github.com/ZiyueWu59/CCA}{https://github.com/ZiyueWu59/CCA}.

%% file: intro.tex
\IEEEPARstart{V}{ideo} Temporal Grounding (VTG) task aims to localize the temporal segment in a video that is semantically aligned with the given language query. It has various applications such as robotic navigation, video surveillance/entertainment, autonomous driving, etc. Recently, many approaches \cite{gao2017tall,yuan2019semantic,zhang2019man,zhang2019cross,zhang2020span,rodriguez2020proposal,wang2020temporally,zhang2020learning,zeng2020dense,mun2020local,liu2021context,liu2022memory,gao2021learning,zhang2020temporal,ning2019attentive,wang2022cross,teng2021regularized,wang2021weakly,zhang2021multi,tang2021frame} have been proposed for the VTG task. Due to the large cross-modal gap between video and text in the VTG task, existing approaches mainly focus on improving the accuracy of localization by designing complicate cross-modal interaction operations. Although they have achieved excellent results on several public datasets, only a few take notice of the test-time cost. In fact, the test-time is an essential metric in many practical applications that need an efficient mechanism to respond quickly when receiving a natural language query. For instance, when enjoying intelligent robot service, the customer may feel impatient or even angry due to slow response and may not employ similar services in the future. Moreover, the response time is more critical in autonomous driving, which always means whether it is safe or not.
Recently, fast video temporal grounding (FVTG) \cite{gao2021fast} is proposed for accurate temporal localization and a efficient test process. Note that the current VTG pipeline can be divided into three components: video encoder, text encoder, and cross-modal interaction module. For the video/text encoders, to obtain more effective information, some traditional modules are widely adopted in most VTG methods for encoding different modality information, \eg, I3D \cite{carreira2017quo} C3D \cite{tran2015learning} for visual encoding and BiLSTM \cite{hochreiter1997long}, GRU \cite{chung2014empirical} for text encoding.
\begin{figure}[!t]
	\centering
	\includegraphics[width=1.0\linewidth]{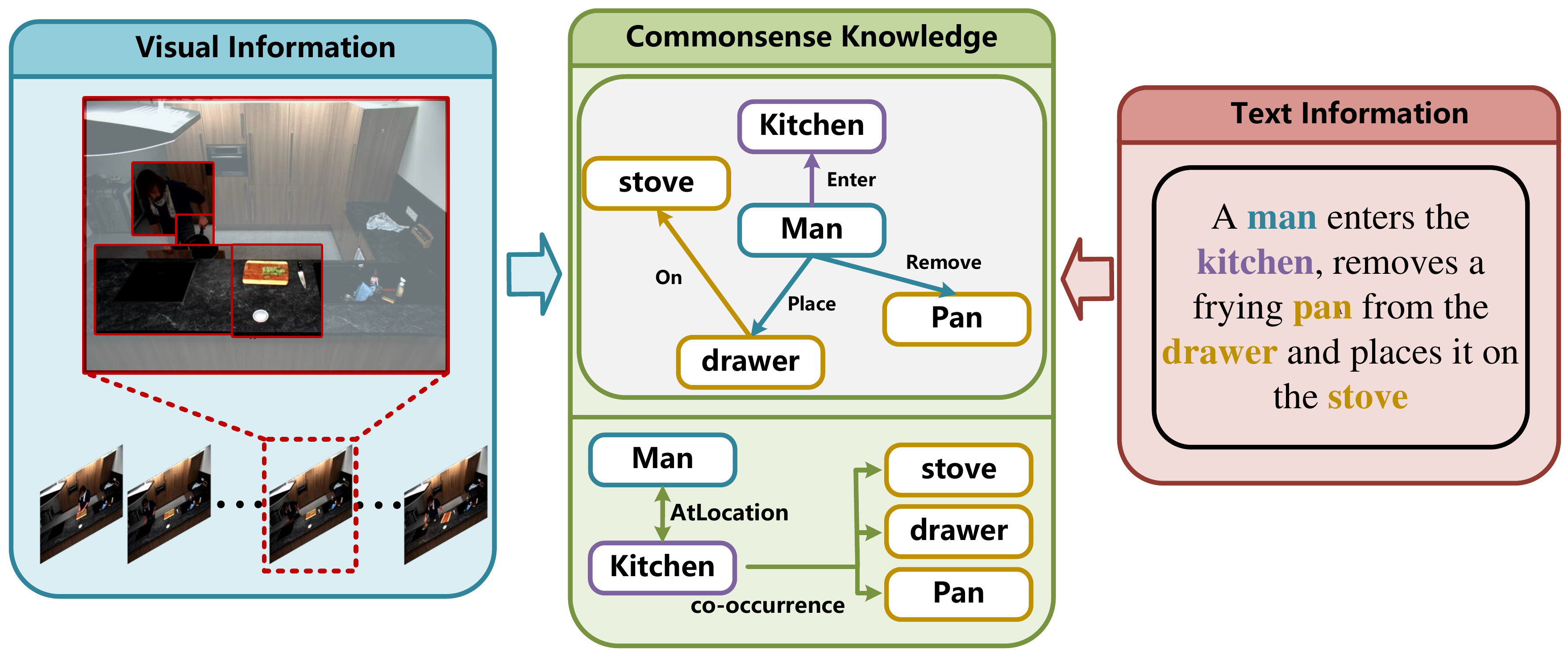}
	\caption{Illustration of Commonsense Knowledge, which is expressed by core semantic concepts and the correlations between visual and text information. It is obvious that the co-occurrence of ``kitchen'', ``pan'' and ``stove'' consists of a kind of commonsense knowledge beacuse they are more likely appear in video or text.}
	\label{fig:intro}
\end{figure}

The most significant component of the VTG is the interaction between different modalities. As a result, existing methods utilized
attention \cite{wang2021structured} or transformer \cite{zhang2021multi}, graph neural networks \cite{zhang2020span, zhang2019cross} and temporal adjacent networks \cite{zhang2020learning} to conduct interaction. Although  bringing rich cross-modal interaction information, this module always consumes the majority of the test-time due to complex feature matrix interaction operation \cite{mun2020local,yuan2019semantic,zeng2020dense} or transformations \cite{ghosh-etal-2019-excl}. Different from the above approaches, to calculate the similarity scores between video moments and texts, common space is utilized in FVTG~\cite{gao2021fast}, where the efficient vector operations like dot production between different modality features are conducted. As a result, the common space-based approaches can achieve a significant test speed. For example, Gao and Xu~\cite{gao2021fast} propose a fast video moment retrieval approach, which achieves not only about 35 times faster than the state-of-the-arts in the cross-modal learning process but also favorable results by employing a fine-grained semantic distilling framework. However, such common space-based methods still have a problem restricting their localization accuracy. That is, they cannot establish the interaction process explicitly between two different modalities, which leads to the features learned in common space hardly capturing the complex relation between them.

Intuitively, without a well-designed cross-modal interaction, it is difficult to ground a textual query onto the video effectively since the information deficiency is brought by inadequate interactions between visual and textual modalities. This deficiency further leads to the lack of learning discrimination of different modality features. Therefore, how to effectively mine the complex cross-modal relation while still maintaining efficient moment-text alignment in the common space is a key issue in FVTG task.
We notice that humans can accurately capture this kind of complex cross-modal relationship in the real world. The reason is that humans have the ability to comprehend and connect the video with the text by using their commonsense knowledge of the world learned through experience, which can be expressed by core semantic concepts and the correlations between them. As shown in \figref{intro}, given a query sentence, ``A man enters the kitchen, removes a frying pan from the drawer and places it on the stove''. When the word ``kitchen'' appears, the words ``pan'' and ``stove'' are more likely to appear in the text. Moreover, seeing such words in a sentence also brings people's minds to visualize objects and their associations with related entities. \eg, when seeing the word ``pan'', it always reminds us to imagine the visual appearance of  ``kitchen'', ``stove'' and other similar words. That is, the co-occurrence of ``kitchen'', ``pan'', and ``stove'' consists of a kind of commonsense knowledge in both visual and textual modalities. Until now, such commonsense knowledge has not been studied and exploited for the VTG task. 

Motivated by the above observation, we propose a Commonsense-aware Cross-modal Alignment framework (CCA) for the fast video temporal grounding task, which incorporates commonsense-guided visual and text representations into a complementary common space and learns efficient commonsense-aware cross-modal alignment. 
To obtain appropriate commonsense knowledge concepts, we utilize the natural language queries to perform statistics and select a set of representative words as concepts. Besides, the relations between these concepts are also considered as a graph, and a concept extractor is employed to obtain commonsense concept features.
After that, a commonsense-aware interaction module is designed on both moment and text sides, which can obtain commonsense-guided visual and text features by adaptively and attentively using concept features to bridge the cross-modal gap and make up for the information deficiency in further common space learning. The learned concept representation could be pre-extracted and stored in an offline manner, which does not affect the test time. To maintain the original semantic information of textual queries, a cross-modal complementary common space is learned to obtain matching scores for all moment proposals and rank them to select the best matching ones. Experiments on two benchmarks show that our proposed framework outperforms other state-of-the-art competitors on overall performance comparison and microscope analyses. 

The main contributions of this paper are summarized as follows:

(1) To the best of our knowledge, we are among the first to exploit commonsense information for fast VTG task without adding additional cross-modal interactions while achieving favorable performance. 

(2) A commonsense-aware interaction module is designed to efficiently leverage the extracted commonsense information, which could serve as a bridge between visual and text modalities. Here, two concept-guided attention modules are utilized to enhance visual and text features for cross-modal alignment, respectively. 

(3) Extensive experiments are conducted on two challenging datasets: TACoS and ActivityNet Captions. The experimental results demonstrate that our proposed method performs superior to the state-of-the-art methods and almost has the highest speed.

%% file: rw.tex
\subsection{Video Temporal Grounding}
Video Temporal Grounding task aims to localize the most relevant video segment corresponding to the given sentence query. VTG has been gaining popularity in the past few years due to its enormous potential applications in video comprehension, to name a few \cite{chen2020rethinking, gao2017unified_video_recommendation, gao2019graph,gao2017deep_relative_tracking,zhang2020span,gao2020ci,gao2020unsupervised}. Existing state-of-the-art VTG methods can be divided into three categories: proposal-based methods and proposal-free methods.

{\noindent \bf{Proposal-based Methods. }} Most Proposal-based methods \cite{gao2017tall,yuan2019semantic,zhang2019man,zhang2020learning,zhang2019cross,ge2019mac, xiao2021boundary, gao2021relation, zeng2021multi, liu2022exploring} are based on several well-designed dense sample strategies, which obtain a set of video segments as candidate proposals and rank them according to the similarity scores calculated between the proposals and the query to select the best matching pairs. Sliding windows is widely used in previous work like CTRL\cite{gao2017tall}, MCN\cite{anne2017localizing}, ACRN\cite{liu2018attentive} and ACL\cite{ge2019mac}. Gao \etal \cite{gao2017tall} first conceive this problem and propose a Cross-modal Temporal Regression Localizer (CTRL) framework by utilizing sliding windows to generate temporal proposals for further fusion with extracted query feature. To improve the quality of the generated moment proposals, Yuan \etal \cite{yuan2019semantic} propose a Semantic Conditioned Dynamic Modulation (SCDM), which can dynamically adjust the temporal convolution according to the query semantics, and integrate the query with visual representations for correlating the sentence-related video contents. Xiao \etal \cite{xiao2021boundary} propose a Boundary Proposal Network (BPNet) by utilizing a third part model to generate high-qualitiy video temporal candidates. To distinguish frame-level features in videos, Liu \etal \cite{liu2022exploring} propose a Motion-Appearance Reasoning Network (MARN), which leverages extracted object features and models their relations for better localization. Besides, rich temporal information is taken into consideration in some work. Zhang \etal \cite{zhang2019man} propose a Moment Alignment Network (MAN) to model complex temporal relations in a video by explicitly establishing relations between different moments and structuring them into a graph for localization in an end-to-end manner. Zhang \etal \cite{zhang2020learning} make full use of the temporal context from different moment proposals by structuring a 2D temporal map to capture the temporal relations between different video moments. For more fine-grained interaction, Zhang \etal \cite{zhang2019cross} propose a Cross-Modal Interaction Network (CMIN), which leverages syntactic structures for fine-grained feature learning and utilizes multi-stage cross-modal interaction to obtain the potential relations between visual and text modalities. In addition, Gao \etal \cite{gao2021relation} regard VTG task as video reading comprehension and propose the Relation-aware Network (RaNet).  Currently, most proposal-based methods are time-consuming due to the large number of proposal-query interactions.

{\noindent \bf{Proposal-free Methods. }} In fact, the impressive performance achieved by proposal-based methods largely depends on the quality of the sampled proposals. Instead of generating numbers of moment proposals as candidates, proposal-free methods \cite{ghosh-etal-2019-excl,mun2020local,zeng2020dense,wang2021structured,liu2022memory,yuan2019find,rodriguez2020proposal, zhao2021cascaded, li2021proposal,chen2020rethinking} directly regress or predict the starting and ending time of the target moment to reduce the extra computational cost brought by the generation of proposal features. Zeng \etal \cite{zeng2020dense} propose a Dense Regression Network (DRN) to regress the distance from each frame to the target moment boundary. Mun \etal \cite{mun2020local} propose a LGI framework to exploit the implicit semantic information by a sequential query attention module from global to local. Also, taking local and global information into account, Liu \etal \cite{liu2021context} propose a Context-aware Biaffine Localizing Network (CBLN) that incorporates local and global contexts into the boundaries with position information for biaffine-based localization. Zhao \etal \cite{zhao2021cascaded} formulate VTG task into a multi-step decision problem and propose the Cascaded Prediction Network (CPN). Besides, Wang \etal \cite{wang2020temporally} aggregate contextual information by obtaining the relations between the current segment and its neighbor segments and propose a Contextual Boundary-aware Prediction (CBP). Liu \etal also consider the contextual information, and propose the Contextual Pyramid Network (CPNet), which models the complex temporal correlation in videos. By addressing VTG task with a span-based QA method, Zhang \etal \cite{zhang2020span} propose a Video Span Localizing Network (VSLNet), which utilizes a query-guided highlighting strategy for matching video segment in the highlighted region. To replenish semantic information, Chen \etal \cite{chen2020rethinking} propose a Graph-FPN with Dense Predictions (GDP) framework to obtain multi-level semantics by using feature pyramid and encoding scene relationships. Considering the off-balance data distribution, Liu \etal \cite{liu2022memory} propose a Memory-Guided Semantic Learning Network (MGSL-Net) to alleviate the forgetting issue by a memory argumentation module. Although these proposal-free methods are more likely to perform effectively and accurately due to the excellent cross-modal interaction and regression operation, most of them regress only one target moment, which does not match the requirements of practical tasks.

With reinforcement learning gradually becoming more popular in recent years, several new reinforcement-learning-based VTG methods have also been proposed. Wu \etal \cite{wu2020tree} formulate a Tree-Structured Policy based Progressive Reinforcement Learning framework (TSP-RPL), which consecutively regulates the predicted boundaries by an iterative refinement process. Besides, the Semantic Matching Reinforcement Learning model (SM-RL) is proposed \cite{wang2019language} to obtain semantic concepts by reinforcement learning and fuse them with context features. Motived by Visual Question Answering tasks(VQA), He \etal \cite{he2019read} regard VTG task as a problem of sequential decision and design a RWM-RL method to regulate the boundaries of predicted results based on its policy.

To improve the localization performance, a lot of existing work tends to apply more complex cross-modal interaction operations. Wang \etal \cite{wang2021structured} propose a Structured Multi-level Interaction Network (SMIN) that utilizes three submodules for the interaction between the visual proposals and different scales of the text representations and uses them iteratively. Zhang \etal \cite{zhang2021multi} propose a Cross-modal Interaction Network in which a multi-stage cross-modal interaction module is designed by complex attention operation. In addition, fine-grained semantic information is gradually explored in VTG task. Chen \etal \cite{chen2020fine} explicitly structure a sentence in three different semantic levels and a graph neural network is used to obtain different level semantic information. Mun \etal and Qu \etal \cite{qu2020fine} consider semantics in a implicit manner, while Ge \etal \cite{ge2019mac}, Chen \etal \cite{chen2020hierarchical} and Jiang \etal \cite{jiang2019cross} in a partial ways. 

Nevertheless, with the improvement of performance, huge and complex architectures inevitably result in higher computational cost during test phase. Therefore, in this paper, a common space is learned in our proposed method for fast video temporal grounding.

\subsection{Fast Video Temporal Grounding}
Recently, fast video temporal grounding (\emph{fast VTG}) has been proposed for more practical applications. VTG task usually requires methods to efficiently localize target video segments in thousands of candidate proposals. In fact, several early algorithms, \eg, common space-learning methods \cite{anne2017localizing,hendricks2018localizing} and skip scanning-based method \cite{hahn2019tripping}, make some contribution to reducing computational cost. Gao \etal \cite{gao2021fast} explore the fast VTG formally and achieve great performance. According to \cite{gao2021fast}, the standard VTG pipeline can be divided into three components. The visual encoder and the text encoder are proved to have little influence in model testing due to the features pre-extracted and stored at the beginning of the test, and cross-modal interaction is the key to reducing the test-time. In \cite{gao2021fast}, a common space learning paradigm is designed to speed up the model. Moreover, a fine-grained semantic distillation framework is utilized to leverage semantic information for improving performance. However, though fine-grained semantic information is used to strengthen the learning of common space, FVMR cannot establish the interaction process explicitly between two different modalities, which leads to the information deficiency in cross-modal relationships. In our proposed method, commonsense knowledge is utilized to obtain bridged visual and text representations, promoting each other in common space learning and making our method perform better.

\subsection{Knowledge Based Visual Related Tasks}
Recently, the usage of knowledge has become a very active field of computer vision by providing external information and efficiently improving the capability of the model. Knowledge information has been widely used in computer vision tasks~\cite{gu2019scene,junyu2019AAAI_TS-GCN,rambhatla2021pursuit,gao2020learning,gao2018watch}. For scene understanding, Gu \etal \cite{gu2019scene} utilize external knowledge and image reconstruction loss to overcome the noisy and missing annotations in datasets. Zheng \etal \cite{zheng2021progressive} obtain knowledge graphs by extracting abundant visual concepts and then combining DNN structures with professional knowledge for scene understanding.
To achieve fine-grained image classification, He \etal \cite{he2021knowledge} propose a Knowledge Graph Representation Fusion (KGRF) framework by using prior knowledge. Besides, Rambhatla \etal \cite{rambhatla2021pursuit} propose a working and semantic memory framework to discover unknown categories when prior knowledge is known. Moreover, a lot of cross-modal comprehension tasks also pay attention to it, such as VQA \cite{wang2017fvqa,marino2021krisp} and image classification \cite{8099493}. Marino \etal \cite{8099493} leverage prior knowledge into graphs and build a Graph Search Neural Network for efficient image classification. They also focus on the required outside knowledge that is not present in the given images and propose the knowledge reasoning with implicit and symbolic representations framework \cite{marino2021krisp} to solve the knowledge-based questions by learning effective implicit symbolic representations. In video-text retrieval, Cao \etal \cite{cao2022visual} explore the visual consensus by structuring them into a graph, and propose a Visual Consensus Modeling (VCM) framework. In image-text matching, Wang \etal \cite{wang2020consensus} propose a Consensus-aware Visual-Semantic Embedding (CVSE) model to mine consensus information in image-text retrieval. 
Compared with CVSE, our proposed CCA method is the first one to leverage commonsense knowledge for temporal modeling in the video temporal grounding task. CCA takes the temporal information in video and text into consideration, while CVSE only fuses consensus features with visual and textual features, respectively. Besides, the commonsense concepts extracted in CCA are from a single dataset where the experiments are conducted, 
while CVSE utilizes large-scale external knowledge to obtain richer knowledge information. Moreover, CVSE categorizes concepts into three types for more detailed information, while CCA selects concepts only based on their frequencies.

%% file: ours.tex
\begin{figure*}[t]
	\centering
	\includegraphics[width=1.0\linewidth]{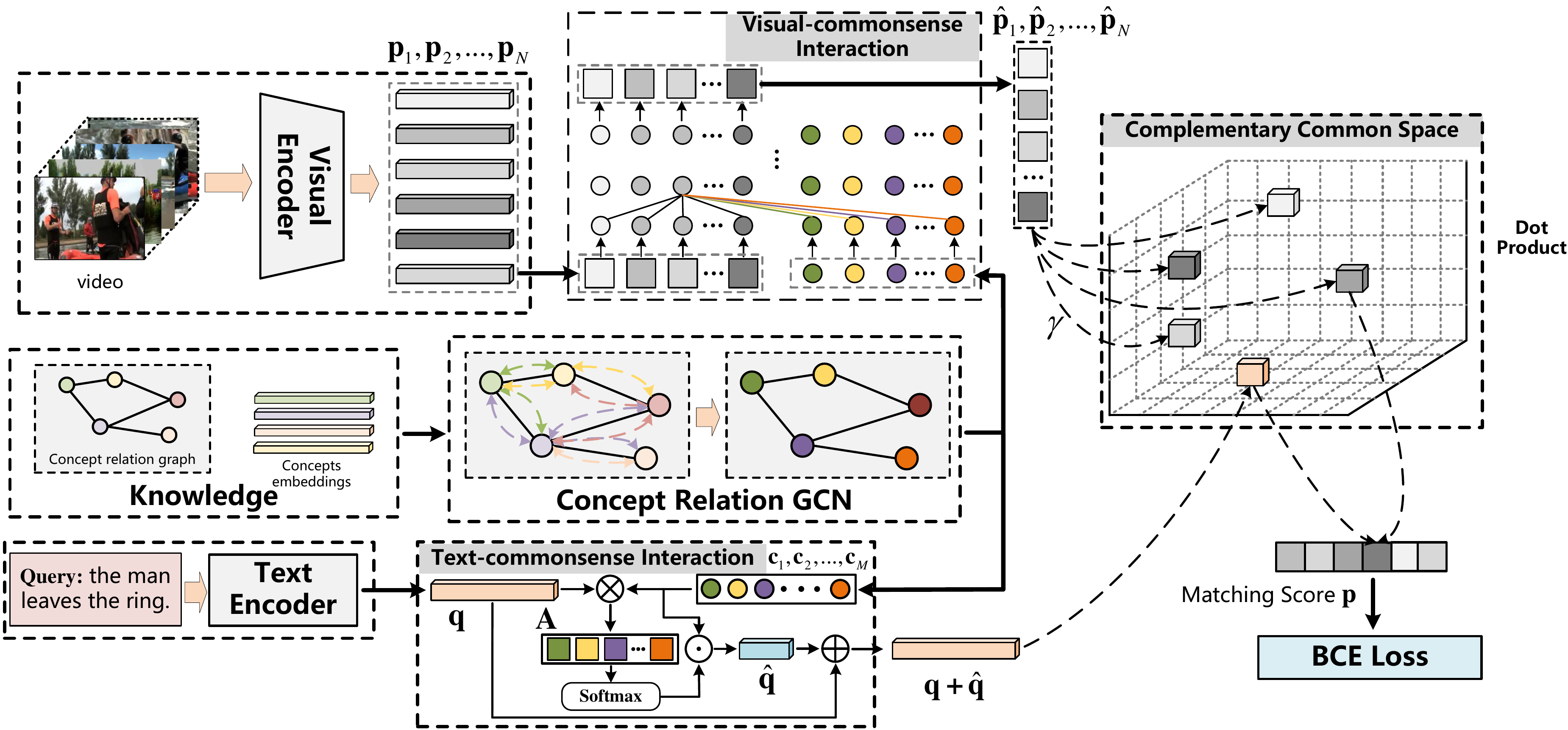}
	\caption{Overview of our proposed CCA framework. CCA mainly consists of three components: multi-modal feature extractor, commonsense-aware interaction module, and complementary common space. We utilize the multi-modal feature extractor to extract visual, text, and commonsense concept features. Then in the commonsense-aware interaction module, two attention-based structures are used to obtain commonsense-guided visual and text features. After that, we map these guided features into two common spaces to calculate the matching scores for each proposal, and we leverage a residual mechanism to obtain final scores. Finally, we rank the scores of all proposals, and a BCE loss is used to optimize the whole framework.}
	\label{fig:framework}
\end{figure*}

In this section, we first introduce the problem formulation of video temporal grounding task and the general scheme of our proposed framework. Then, we present each component of our framework, including multi-modal feature extractor, commonsense-aware interaction module, and complementary common space, as shown in \figref{framework}. Finally, the training and inference settings of our framework are present.

\subsection{Problem Formulation}
Given an untrimmed video $ V $ and a natural language query $ Q $, the goal of video temporal grounding is to localize the video segment $ (t_{s}, t_{e}) $ that is most relevant to the query, where $ t_{s} $ and $ t_{e} $ denote the start and the end time of the target moment. The video is denoted as a sequence of frames $ V=\{v_{i}\}^{T}_{i=1} $, where $ T $ is the number of all the frames, and the corresponding text query is denoted as $ Q=\{w_{j}\}^{L}_{j=1} $ where $ L $ is the length of the query and $ w_{j} $ is the $ j $-th word in the query. 

\subsection{General Scheme} 
Aiming to solve the problem that existing VTG methods always resort to complex and time-consuming cross-modal interaction modules to obtain rich information, 
we propose a commonsense-aware cross-modal interaction framework to make full use of commonsense knowledge. In this way, we can obtain more accurate results on the VTG task without adding additional cross-modal interactions. \figref{framework} illustrates the architecture of our proposed CCA method. Our approach consists of three components: a multi-modal feature extractor, a commonsense-aware interaction module, and a complementary common space. We first utilize three encoders to extract video, query, and commonsense features. Then a commonsense-aware interaction module is designed to generate commonsense-guided visual and textual representations. After that, we leverage two different attention-based structures to adaptively interact the extracted commonsense features with the video and text features, respectively. Finally, a complementary common space is learned to obtain commonsense-guided moment-query matching scores for temporal localization.

\subsection{Multi-modal Feature Extractor}
In the multi-modal feature extractor, three types of encoders are utilized to extract visual, text and commonsense features, respectively.

{\noindent \bf{Visual encoder.}} We firstly generate a set of video moment candidate proposals denoted as $ P=\{p_{n}\}^{N}_{n=1} $, where $ p_{n} $ is the $ n $-th proposal and $ N $ means the number of all generated proposals. Here, we simply adopt the commonly used 2D-temporal proposal generation approach \cite{zhang2020learning}.  Then, a pre-trained CNN model (\eg {I3D, C3D}) is utilized to extract visual features for each proposal as follows:
\begin{equation}
\label{visual encoder}
	\textbf{P}=\textbf{visEncoder}(P)=\{\textbf{p}_{1}, \textbf{p}_{2}, ..., \textbf{p}_{N}\},
\end{equation}

where $ \textbf{P} \in \text{R}^{N \times d^{V}} $ and $ d^{V} $ represents the dimension of the extracted feature. $ \textbf{p}_{n} $ means the visual features of proposal $ p_{n} $. 

{\noindent \bf{Text encoder:}} For the text query, we leverage a Bi-LSTM \cite{hochreiter1997long} to integrate the sequential information of the word list as follows:
\begin{equation}
\label{text encoder}
	\textbf{Q}=BiLSTM(Q)=\{\textbf{q}_{1}, \textbf{q}_{2}, ..., \textbf{q}_{L}\},
\end{equation}

where $ \textbf{Q} \in \text{R}^{L \times d^{Q}} $ and $ d^{Q} $ means the dimension of the extracted query feature. 
We obtain the word-level feature $ \textbf{q}_{i} $, where $ \textbf{q}_{i}=[\overrightarrow{\textbf{q}}_{i};\overleftarrow{\textbf{q}}_{i}] $ through the concatenation of hidden states in both directions of Bi-LSTM, and calculate the sentence-level feature  $ \textbf{q}=[\overrightarrow{\textbf{q}}_{L};\overleftarrow{\textbf{q}}_{1}] $ by concatenating the last hidden state of both forward and backword LSTM.

{\bf{Commonsense Concept Encoder:}} To capture the  high-level relation information as in human reasoning, which is referred to as "commonsense knowledge concept", we choose some high-frequency words to serve as our commonsense concept. Specifically, 
we follow \cite{dong2021dual} to select $ M $ frequent concepts as commonsense concepts in the language corpus.
After getting these concepts, we use Glove-300 \cite{pennington2014glove} to initialize the concepts denoted as $ C=\{c_{1}, c_{2}, ..., c_{M}\} $, where $ M $ is the number of concepts. 
According to \cite{wang2020consensus}, we use both normalized co-occurrence frequency and word embedding similarity to initialize the concept relation graph $ \textbf{G} $. 
Finally, a two-layer Graph Convolution Network (GCN) is employed to extract commonsense concept features as follows:

\begin{equation}
\centering
\label{gcn}
\begin{aligned}
	\textbf{H}^{(l+1)}&=f(\textbf{H}^{(l)}, \textbf{A})=\sigma(\textbf{A}\textbf{H}^{(l)}\textbf{W}^{(l)}), \\
\textbf{A} 	&= \textbf{D}^{-1/2}\textbf{G}\textbf{D}^{-1/2},
\end{aligned}
\end{equation}
where $ \textbf{D} $ represents degree matrix which is used to normalize $ \textbf{G} $. $ \textbf{W}^{l} $ is a learnable weight matrix. $ \sigma $ is a non-linear activation function like $ ReLU $.

After GCN, the node representation produced by the last graph convolutional layer is regarded as commonsense concept features $ \textbf{C}=\{\textbf{c}_{1}, \textbf{c}_{2}, ..., \textbf{c}_{M}\} $, where $ \textbf{C} \in \text{R}^{M \times d^{C}} $ and $ d^{C} $ means the dimension of concept feature.

\subsection{Commonsense-aware Interaction Module}
In traditional approaches, there is no mechanism to explore the commonsense knowledge for the cross-modal bridging in the VTG task. To address this drawback, we construct a commonsense-aware interaction module to achieve the goal by associating the extracted commonsense concept with both visual and textual representations.
After encoding the moment proposals, sentence, and commonsense concepts, two adaptive attention-based structures, visual-commonsense interaction and text-commonsense interaction modules, are used for constructing cross-modal alignment.

{\noindent \bf{Visual-Commonsense Interaction Module.}} Recently, the attention mechanism is widely used in cross-modal learning tasks for modality interaction. However, with the complex temporal relations between different proposals and the high-level relations in commonsense concepts, a simple soft attention mechanism that captures interaction from one specific attention space may not be enough to achieve comprehensive information passing between different modalities. Therefore, a visual-commonsense interaction module is designed for robust relation modeling and better interaction between visual proposals and commonsense concepts. Specifically, we first concatenate moment proposals $ \textbf{P} $ with concept features $ \textbf{C} $ into a unified sequence:
\begin{equation}
		\textbf{F}^{Cat}=Concat(\textbf{P}, \textbf{C})
\end{equation}
where $ \textbf{F}^{Cat} $ means that $ \textbf{P} $ and $ \textbf{C} $ are concatenated in their spatial dimension, $\textbf{F}^{Cat} \in \text{R}^{(N+M) \times d^{V}}$. Specially, $ d^{C} $ is set the same as $ d^{V} $. After that, a multi-head self-attention module is employed to process this long sequence appropriately and efficiently, which could allow our model to jointly attend to the information from different positions and capture richer characteristics from diverse modalities.
\begin{equation}
	\begin{aligned}
	\textbf{A}^{Cat} &= \textbf{F}^{Cat}\textbf{W}^{Q}_{i}(\textbf{F}^{Cat}\textbf{W}^{K}_{i})^\top, \\
	head_{i}&=softmax(\frac{\textbf{A}^{Cat}}{\sqrt{d^{avg}}})\textbf{F}^{Cat}\textbf{W}^{V}_{i},\\
	\textbf{F}^{mul}&=Concat(head_{1}, head_{2}, ..., head_{n})\textbf{W}^{mul}, \\
	\end{aligned}
\end{equation}
where $ \textbf{W}^{Q}_{i}, \textbf{W}^{K}_{i}$ and $\textbf{W}^{K}_{i} \in \text{R}^{d^{avg} \times d^{V}} $ are learnable weight matrices in the $ i $-th head of the multi-head self-attention, and $ d^{avg} = (N+M) / n $, $ n $ is the number of the parallel heads. $ \textbf{F}^{mul} $ represents the output of multi-head structure, and $\textbf{W}^{mul} \in \text{R}^{d^{avg}} $ is also a learnable weight matrix.

After exchanging information between our concatenated feature $ \textbf{F}^{cat} $ at all spatial positions and obtaining $ \textbf{F}^{mul} $, we further leverage two linear layers and a layer normalization operation to generate the high-level multi-head feature, which serves as a residue to $ \textbf{F}^{mul} $. Finally, we can obtain visual-commonsense feature $ \textbf{F}^{cg} $ as follows:
\begin{equation}
\begin{aligned}
\textbf{F}^{cg} = \textbf{F}^{mul} + layerNorm(\textbf{W}^{q}(ReLU(\textbf{W}^{p}\textbf{F}^{mul}) + \textbf{b}_{p}) + \textbf{b}_{q}),
\end{aligned}
\end{equation}
where $ \textbf{W}^{p} $, $ \textbf{W}^{q} $, $ \textbf{b}_{p} $ and $ \textbf{b}_{q} $ are the learnable parameters of two linear layers.

After normalizing the visual-commonsense feature, we select the top-$ N $ dimension of $ \textbf{F}^{cg} $ to generate commonsense-guided moment proposal features, denoted as $ \hat{\textbf{P}} $:
\begin{equation}
\begin{aligned}
\{\hat{\textbf{p}}_{1}, \hat{\textbf{p}}_{2}, ..., \hat{\textbf{p}}_{N}\}=\hat{\textbf{P}} = Norm(\textbf{F}^{cg})[:N,:],
\end{aligned}
\end{equation}
here, $Norm(\cdot)$ means L2 normalization.

{\noindent \bf{Text-Commonsense interaction module:}} Different from the visual side that has a large number of moment proposals, there is only one text query for a VTG task. Considering different modalities with their specific contents and relation patterns, we design a text-commonsense interaction module to obtain commonsense-guided text representation. Concretely, we first calculate the attention weights $ \textbf{A} $, which represents the pair-wise relations between text and concepts:
\begin{equation}
\begin{aligned}
\textbf{A}=\textbf{q}\textbf{W}^{Q}(\textbf{C}\textbf{W}^{K})^\top,
\end{aligned}
\end{equation}
where $ \textbf{W}^{Q} \in \text{R}^{d^{Q}} $ and $ \textbf{W}^{K} \in \text{R}^{d^{C}} $ are learnable parameter matrices. Then, a softmax function is employed to obtain commonsense-guided query representation:
\begin{equation}
\begin{aligned}
	ConAttn(\textbf{q}, \textbf{C}, \textbf{C}) =softmax(\frac{\textbf{A}}{\sqrt{d^{avg}}})\textbf{C}\textbf{W}^{V},
\end{aligned}
\end{equation}

The obtained representation is then normalized as the final query representation, denoted as $ \hat{\textbf{q}} $, to calculate matching scores in further common space learning:
\begin{equation}
\begin{aligned}
\label{t-c}
\hat{\textbf{q}} = Norm(ConAttn(\textbf{q}, \textbf{C}, \textbf{C})).
\end{aligned}
\end{equation}

\subsection{Complementary Common Space} The performance of \cite{gao2021fast} shows that replacing complex cross-modal interaction with common space learning has a significant effect on reducing test-time cost. Nonetheless, common space learning can not explicitly establish the interaction process between two different modalities, leading to the features learned in common space hardly capturing the complex relations between these modalities.

To solve this problem, commonsense knowledge has been learned and fused with visual and text information, which can efficiently enhance the discriminative ability of different modality features. 
Nevertheless, the commonsense-guide information may lose the global information of the query sentence since commonsense consists of separate concepts, which leads to incomplete VTG learning. 
Therefore, in this section, we design a complementary common space to effectively compute matching scores while maintaining the original semantic information of the video and query. 
To reduce the number of parameters and the cost of time, we simply adopt two feature transformation modules, $ \phi_{1} $, $ \phi_{2} $  to project commmonsense-guided visual features for both the original query features and commonsense-guided query features, respectively. Then the final matching scores can be calculated adaptively as follows:

\begin{equation}
\begin{aligned}
\label{matching scores}
&m_{i}=\phi_{1}(\hat{\textbf{p}}_{i})^\top {\textbf{q}}, \\
&n_{i} =\phi_{2}(\hat{\textbf{p}}_{i})^\top \hat{\textbf{q}}, \\
&a_{i} = \gamma m_{i} + (1-\gamma)n_{i}, \\
\end{aligned}
\end{equation}

where $ \phi_{1} $ and $ \phi_{2} $ are MLPs. For each proposal feature $ \hat{\textbf{p}}_{i} $, $ m_{i} $ represents the matching score between $ \hat{\textbf{p}}_{i} $ and original query feature $ \textbf{q} $, and $ n_{i} $ represents the matching score between $ \hat{\textbf{p}}_{i} $ and commonsense-guided query feature $ \hat{\textbf{q}} $. Then, a learnable parameter, $ \gamma $ is utilized to adaptively balance the weight of the two types of matching score. Finally, $ a_{i} $ means the matching score of the $ i $-th moment proposal for temporal localization.

\subsection{Training and Inference}
{\noindent \bf{Training:}} 
With the above calculated matching scores, a binary cross entropy loss is used to optimize our model as follows:
\begin{equation}
\begin{aligned}
\mathcal{L}(y_{i}, a_{i})=-\frac{1}{N}\sum_{i=1}^{N}y_{i}\log a_{i} + (1-y_{i})\log (1-a_{i}),
\end{aligned}
\end{equation}
where the soft label $ y_{i} $ is generated by thresholding the overlap ratio between the $i$-th moment proposal and the groundtruth temporal segment. 

{\noindent \bf{Inference:}} During testing, we employ the matching scores $ a $ of each proposal for video temporal grounding. The  time cost of our proposed CCA method only consists of the calculation of text encoder, text-commonsense interaction and matching scores since the moment proposal features can be pre-calculated and stored in a gallery database.

%% file: expr.tex
In this section, we evaluate our approach for video temporal grounding task on two public datasets: ActivityNet Caption \cite{krishna2017dense} and TACoS \cite{regneri2013grounding}.

\begin{table*}[t]
	
	\setlength{\tabcolsep}{6pt}
	\caption{Speed-accuracy analysis on two datasets. \textbf{TE}: time cost of query (Text) Embedding generation. \textbf{CML}: time cost of the Cross-Modal Learning for VMR. \textbf{ALL}: The total time cost of TE and CML. We report the accuracy (\textbf{ACC}) of R@1, IoU=0.5 and the sum of the accuracy (\textbf{sumACC}, the sum of the accuracy of IoU=0.3 and IoU=0.5) for comparison. }\label{tab:speed_accuracy}
	\begin{center}
		{	
			\begin{tabular}{l | c c c c c|| c c c c c}
				\hline
				\multicolumn{1}{l|}{\multirow{2}{*}{Methods}} & \multicolumn{5}{c||}{{TACos}} & \multicolumn{5}{c}{{ANetCap}}  \\
				
				\cline{2-11}
				& TE & CML & ALL & ACC & sumACC & TE & CML & ALL & ACC & sumACC \\
				\hline
				
				TMLGA  &\textbf{1.14} &11.37 & 12.51 & 21.65 & 46.19 & \textbf{1.24} &8.97 &10.21 & 33.04 & 84.32\\
				VSLNet & 3.58 & 5.02 & 8.59 & 24.27 & 53.88 & 3.87 & 4.86 &8.74 & 43.22 & 106.38 \\
				LGI    & - & - & - & - & - &\underline{1.53} &7.03 & 8.56 & 41.51 & 100.03\\
				DRN    & 4.67 & 22.13 & 26.81 & 23.17 & - & 4.86 & 18.46 & 23.32 & \underline{45.45} & - \\ \hline
				CTRL   & 4.32 & 534.23 & 538.55 & 13.30 & 31.62 & 4.75 & 398.25 & 403.0 & 29.01 & - \\
				SCDM   & 3.65 & 780.0 & 783.65 & 21.17 & 47.28 & 3.27 & 359.76 & 363.03 & 36.75 & 91.55 \\
				CBP    & 3.17 & 2659.01 & 2662.18 & 24.79 & 52.10 & 2.44 & 522.65 & 525.09 & 35.76 &  90.06 \\
				2D-TAN & \underline{1.72} & 135.84 & 137.56 & 25.32 & 62.61 & 1.69 & 80.35 &403.1 & 44.51 & 103.96 \\
				FVMR   & 3.51 &\textbf{0.14} & \underline{3.65} & \underline{29.12} & \underline{70.60} & 3.14 & \textbf{0.09} & \underline{3.23} & 45.00 & \underline{106.60} \\				
				\hline
				\textbf{Ours} & 2.33 & \underline{0.29} & \textbf{2.62} & \textbf{32.83} & \textbf{78.13} & 2.80 & \underline{0.30} & \textbf{3.10} & \textbf{46.19} & \textbf{106.77} \\
				\hline
			\end{tabular}
		}\label{tab:speed comparison}
	\end{center}
\end{table*}

\subsection{Datasets and Evaluation}
\Paragraph{TACoS:} The TACoS dataset is widely used on VTG task, and it is collected by Regneri \etal~\cite{regneri2013grounding}, which consists of 127 videos from different cooking scenarios. The video's average duration is about 4.79 minutes. TACoS is a challenging dataset for VTG due to the multi-level activities contained by the query sentences in the dataset. Following the standard split \cite{gao2017tall}, TACoS has 10146, 4586, and 4083 moment-query pairs for training, validation, and testing, respectively.

\Paragraph{ActivityNet Captions:} The ActivityNet Captions dataset is a popular benchmark dataset containing around 20K  videos with 100K annotations. Currently, it is the largest dataset for video moment retrieval task. Following~\cite{zhang2020learning, zeng2020dense}, we use the first validation set for validation and the second validation set for testing. The total length of all videos is over 648 hours, and the videos are associated with more than 200 types of daily activities. The videos contain 3.65 moment-sentence pairs on average, and the average length of the descriptions is 13.48 words.

\Paragraph{Metrics:} Following previous work ~\cite{gao2017tall,zeng2020dense}, the evaluation metrics ``R@n,IoU=m'' is utilized to evaluate the ability of our approach. ``R@n, IoU=m'' represents the percentage of at least one of the top-n predicted segments which have Intersection over Union (IoU) larger than m. Specifically, we set n$\in$\{1, 5\} and m$\in$\{0.1, 0.3, 0.5, 0.7\} for ActivityNet Caption dataset and TACoS dataset\label{key}.

\subsection{Implement Details} 
\Paragraph{Feature Extractor.} The same visual features (\ie VGG \cite{simonyan2014very}, C3D \cite{tran2015learning} and I3D \cite{carreira2017quo}) as previous approaches are employed to produce a fair and detailed comparison to reduce the influence of different visual encoders. We adopt VGG, C3D and I3D to make comparison on TACoS, and the latter two on ActivityNet Captions.  

\Paragraph{Concept Relation Graph.} 
For concept selection, in the training set, we firstly obtain the frequencies of each word and select those greater than or equal to 3 on TACoS and 5 on ActivityNet Captions. We also follow~\cite{wang2020consensus} to expand the commonsense concepts. Then, we perform a statistical computation to obtain the correlations of the concepts which belong to training set to generate the concept relation graph. For the concepts that are unseen in the training set, we calculate their cosine similarities with other concepts, and then make similarities as the weight in relation graph. Finally, we obtain 624 commonsense concepts on TACoS and 3152 on ActivityNet Captions.

\Paragraph{Architecture settings.} As for text encoding, we set the max length of word sequence to 30, and utilize the pre-trained 300-dim Glove embeddings \cite{pennington2014glove} to initialize the words in the query and the consensus concepts. Besides, a two-layer bi-directional LSTM is adopted with 512 hidden state dimensions. We use a two-layer Graph Convolution Network (GCN) with 512 and 1024 embedding dimensions for concept feature extraction. The feature dimensions $ d^{V} $, $ d^{Q} $ and $ d^{C} $ are all set to 512.

\Paragraph{Training and Inference settings.} We follow \cite{zhang2020learning} to generate candidate moment proposals by adopting sliding windows to random select $ N $ consecutive clips, which is structured as 2D feature map. The window size $ N $ is set to 128 for TACoS and 64 for ActivityNet Captions, respectively. 
Then, non-maximum supression (NMS) is used on our predicted temporal segments.  The NMS threshold is set to 0.49 for all experiments. Our proposed method is trained by an Adam optimizer \cite{da2014method} with a learning rate of 0.0001. Our model is trained for 50 epochs on TACoS dataset and 30 epochs on ActivityNet Captions dataset, and the batch size is set to 64. All our experiments are implemented in PyTorch toolkit with 4 NVIDIA Geforce RTX 3090 GPUs.

\subsection{Comparison with State-of-the-art Methods} We compared our proposed CCA approach with the following state-of-the-art baselines on two banchmark datasets: 
\begin{itemize}
	\item \textbf{Proposal-based Methods}: CTRL\cite{gao2017tall}, MCN\cite{anne2017localizing}, MAN\cite{zhang2019man}, SCDM\cite{yuan2019semantic}, SAP\cite{chen2019semantic}, TGN\cite{chen2018temporally}, ACRN\cite{liu2018attentive}, QSPN\cite{xu2019multilevel}, CMIN\cite{zhang2019cross}, FIAN\cite{qu2020fine}, 2D-TAN\cite{zhang2020learning}.
\end{itemize}
\begin{itemize}
	\item \textbf{Proposal-free Methods}: ABRL\cite{yuan2019find}, TMLGA\cite{rodriguez2020proposal}, LGI\cite{mun2020local}, DRN\cite{zeng2020dense}, VSLNet\cite{zhang2020span}, DEBUG\cite{lu2019debug}, ExCL\cite{ghosh-etal-2019-excl}, CBP\cite{wang2020temporally}, CBLN\cite{liu2021context}, MGSL-Net\cite{liu2022memory}.
\end{itemize}
\begin{itemize}
	\item \textbf{Reinforcement-learning-based Methods}: RWM-RL\cite{he2019read}, SM-RL\cite{wang2019language}, TSP-RPL\cite{wu2020tree}, TripNet\cite{hahn2019tripping}.
\end{itemize}

In the following, the best performance is highlighted in \textbf{bold} and the second-best \underline{underline}.

\begin{table*}[t]
	\caption{Comparison results on TACoS. }
		\begin{center}
			{
				\begin{tabular}{c|ccccc|ccccc}
					\hline
					\multirow{2}*{Method} & \multicolumn{5}{c|}{R@1} & \multicolumn{5}{c}{R@5} \\
					\cline{2-11}
					& IoU=0.1 & IoU=0.3 & IoU=0.5 & IoU=0.7 & sumACC & IoU=0.1 & IoU=0.3 & IoU=0.5 & IoU=0.7 & sumACC\\ \hline 
					\multicolumn{11}{c}{VGG features} \\
					\hline
					MCN      & 14.42 & - & 5.58 & - & - & 37.35 & - & 10.33 & - & - \\
					SM-RL    & 26.51 & \underline{20.25} & 15.95 & - & \underline{36.20} & 50.01 & \underline{38.47} & 27.84 & - & \underline{66.31} \\
					SAP      & \underline{31.15} & - & \underline{18.24} & - & - & \underline{53.51} & - & \underline{28.11} & - & - \\
					\hline
					\textbf{Ours} & \textbf{54.55} & \textbf{43.25} & \textbf{30.52} & \textbf{17.02} & \textbf{73.77} & \textbf{78.38} & \textbf{65.97} & \textbf{52.78} & \textbf{31.88} & \textbf{118.75} \\ \hline
					\hline
					\multicolumn{11}{c}{C3D features} \\
					\hline
					TGN      & 41.87 & 21.77 & 18.90 & 11.88 & 40.67 & 53.40 & 39.06 & 31.02 & 15.26 & 70.08 \\
					ACRN     &   -   & 19.52 & 14.62 &   -   & 34.14 &   -   & 34.97 & 24.88 &   -   & 59.85 \\ 
					DEBUG    & 41.15 & 23.45 &   -   &   -   &   -   &   -   &   -   &   -   &   -   &   -   \\
					DRN      &   -   &   -   & 23.17 &   -   &   -   &   -   &   -   & 33.36 &   -   &   -   \\ 
					CTRL     & 24.32 & 18.32 & 13.30 & - & 31.62 &   48.73   & 36.69 & 25.42 & -  & 62.11 \\
					QSPN     & 25.31 & 20.15 & 15.23 & -  & 35.38 &  53.21 & 36.72 & 25.30 & -  & 62.02\\
					ACL      & 31.64 & 24.17 & 20.01  & -  & 44.18 &  57.85 & 42.15 & 30.66 & -  & 72.81 \\
					SCDM     &  -    & 26.11 & 21.17  & -  & 47.28 &  - & 40.16 & 32.18 & -  & 72.34 \\
					CBP      &   -   & 27.31 & 24.79 & -  & 52.10 &  - & 43.64 & 37.40 & -  & 81.04 \\
					2D-TAN   & 47.59 & 37.29 & 25.32 & - & 62.61 & 70.31 & 57.81 & 45.04 & -  & 102.85 \\
					FIAN     & 39.55 & 33.87 & 28.58 & - & 62.45 & 56.14 & 47.76 & 39.16 &  -  & 86.92 \\ 
					CBLN     & 49.16 & 38.98 & 27.65& -  & 66.63 & 73.12 & 59.96 & 46.24 & -  & 106.20 \\
					CMIN     &   -   & 24.64 & 18.05 & - & 42.69 & - & 38.46 & 27.02 & - & 65.48 \\ 
					TripNet  &   -   & 23.95 & 19.17 & 9.52 & 43.12 & - & - & - & - & - \\
					ABLR     & 34.70 & 19.50 & 9.40 & - & 28.90 & - & - & - & - & - \\ 
					BPNet    &   -   & 25.96 & 20.96 & - & 46.92 & - & - & - & - & - \\
					MGSL-Net &   -   & \underline{42.54} & \underline{32.27} & -  & \underline{74.81} & - &  63.39 & \underline{50.13} & - & 113.52 \\
					FVMR     & \underline{53.12} & 41.48 & 29.12 & \underline{16.35} & 70.60 & \textbf{78.12} &  \textbf{64.53} & 50.00 & \underline{30.15} & \underline{114.53} \\ \hline
					\textbf{Ours} & \textbf{56.00} & \textbf{45.30} & \textbf{32.83} & \textbf{18.07} & \textbf{78.13} & \underline{76.60} & \underline{64.38} & \textbf{52.68} & \textbf{33.10} & \textbf{117.06} \\ \hline
					\hline
					\multicolumn{11}{c}{I3D features} \\
					\hline
					ExCL     & - & \textbf{45.50} & \underline{28.00} & 13.80 & \underline{73.50} & - & - & - & - & - \\
					TMLGA    & - & 24.54 & 21.65 & \underline{16.46} & 46.19 & - & - & - & - & - \\
					VSLNet   & - & 29.61 & 24.27 & - & 53.88 & - & - & - & - & - \\ \hline
					\textbf{Ours} & \textbf{54.38} & \underline{43.30} & \textbf{31.08} & \textbf{18.05} & \textbf{74.38} & \textbf{77.68} & \textbf{64.03} & \textbf{52.78} & \textbf{32.83} & \textbf{116.47} \\ \hline
				\end{tabular}
			}\label{tab:tacos results}
		\end{center}
		\vspace{-2mm}
	\end{table*}

{\noindent \bf{Overall Speed-Accuracy Analysis.}} Considering that fast VTG task pays the same attention to the speed as the accuracy, following \cite{gao2021fast} we evaluate the time cost of Text Encoding (TE) for query embedding generation and Cross-Modal Learning for moment localization (CML). ~\tabref{speed comparison} shows the performance. Besides, we also calculate the sum of the accuracy in terms of ``IoU=0.3'' and ``IoU=0.5'', named sumACC to evaluate the whole performance of each model. Due to the learning of consensus knowledge and the simple but efficient dot production between different modal feature vectors in learned common space, our proposed method achieves great performance with both high speed and efficiency. Obviously, TE has no effect on the test-time, and each method spends a similar time ( $\sim3ms $) on TE module because of the limited capability of text encoders like LSTM. However, a vast difference appears in the time cost of CML between different approaches. For the CML, we can find that our proposed method is at least fifty times faster than state-of-the-arts that do not learn common space. Compared with FVMR, with the similar time cost, our proposed method outperforms on ``R@1, IoU=0.5'' by gains of $ 3.71\% $ on TACoS dataset and $ 1.19\% $ on ActivityNet Captions dataset. According to sumACC, we can find that our proposed CCA outperforms other state-of-the-art methods by gains of at least $ 7.53\% $ on TACoS and $ 0.17\% $ on ActivtyNet Captions. It demonstrates that the learning of consensus knowledge can obtain more discriminative features for better performance. Though these proposal-free approaches such as LGI, VSLNet, and TMLGA also achieve favorable performance with low computational expenses, our proposed method still outperforms them by gains of $ 4.68\% $, $ 2.97\% $, $ 13.15\% $ on ActivityNet Captions, respectively. Moreover, most proposal-free methods can only regress one temporal
location for VTG, which is not suitable in practical applications. The above comparison illustrates that our method has significant speed and accuracy advantages.

{\noindent \bf{Results on TACoS and ActivityNet Captions. }} We compare the performance of our proposed method against extensive video temporal grounding models on two benchmark datasets. As shown in~\tabref{tacos results}, we can observe that our method performs better than other methods in most metrics. On the TACoS dataset, our proposed CCA achieves significant performance on all the three types of visual features. Compared with FVMR, thought our method achieves $ (1.52\%, 0.15\%) $ lower than FVMR on metric ``R@5, IoU=\{0.1, 0.3\}'' on TACoS, it outperforms FVMR in all other metrics, especially on the metrics ``R@1, IoU=\{0.5, 0.7\}'' and ``R@5, IoU=\{0.5, 0.7\}'' by gains of $ (3.71\%, 1.72\%) $ and $ (2.68\%, 2.95\%) $ respectively. Note that IoU=0.7 is a more crucial criterion to determine whether a VTG model is accurate or not. The comparison of performance on ToU=0.7 shows that our method can predict results with higher quality. Moreover, as shown in \tabref{acnet results} our proposed CCA also surpasses FVMR on ActivityNet Captions by $ (0.98\%, 1.55\%) $ in terms of ``R@1, IoU=\{0.5, 0.7\}''. ActivtyNet Captions has larger scales than the TACoS dataset. The results indicate that our method also performs well in a more complex visual-text environment.

Then, we compare our model with much more VTG methods in more detail. Firstly, we compare CCA with previous proposal-based methods: CTRL, MCN, MAN, ACL, TGN, QSPN, CMIN, CBP, SCDM and 2D-TAN. From the results in \tabref{tacos results} and \tabref{acnet results}, we observe that our CCA achieves great performance compared with the aforementioned methods on most of the metrics. Part of previous work ignores the temporal context information as well as making inadequate cross-modal interaction. Meanwhile, our CCA model replaces complex cross-modal interaction with commonsense knowledge, which makes the generated features more discriminative, and saves much more time cost. The experimental results demonstrate the effectiveness of CCA in capturing rich cross-modal information and characterizing the complex associative patterns between different modalities.  

Moreover, we compare our method with previous proposal-free methods: TGN, CMIN, CBP, SCDM, DRN, LGI, CBLN, and MGSL-Net. Due to the fine-grained interaction and the regression product, the mentioned methods have achieved great results in recent years. DRN mainly regresses the distances from each frame to the temporal boundaries, LGI combines local and global information, while MGSL-Net pays more attention to the off-balance data distribution. 
Compared with them, our proposed CCA method achieves better performance. On ActivityNet Captions, we outperforms LGI by gains of $ (2.06\%, 4.68\%, 5.80\%) $ in terms of ``R@1, IoU=\{0.3, 0.5, 0.7\}", and outperforms DRN by gains of $ (0.74\%, 4.51\%) $ in terms of ``R@1, IoU=\{0.5, 0.7\}''. Compared with MGSL-Net on TACoS, our method outperforms it by gains of $ (2.76\%, 0.56\%, 0.99\%, 2.55\%) $ on the metrics ``R@\{1, 5\}, IoU=\{0.3, 0.5\}''. Besides, it also surpasses the recent work CBLN with an average $ 6.11\% $ improvement on the metrics ``R@1, IoU=\{0.1, 0.3, 0.5\}''. Note that other state-of-the-art approaches hardly show their results on the metric ``IoU=0.7'' on TACoS, and ``IoU=0.7'' is a stricter standard to define whether a localized moment is correct. It indicates that our method can localize the moment with higher quality. Our CCA obtains more accurate results because the extracted commonsense knowledge bridges different modalities and makes the model have the ability to comprehend complex cross-modal relationships. In fact, the simple but effective use of commonsense knowledge replaces large and repetitive cross-modal interaction and reduces the time cost of capturing fine-grained information. It validates that CCA can efficiently and effectively localize the target moment boundary.

\begin{table}[htbp]
	\caption{Comparison results on ActivityNet Captions. }
	\vspace{-2mm}
	\resizebox{85mm}{!}{%
		\begin{tabular}{c|ccc|ccc}
			\hline
			\multirow{2}*{Method}& \multicolumn{3}{c|}{R@1} & \multicolumn{3}{c}{R@5} \\
			\cline{2-7}
			& IoU=0.3 & IoU=0.5 & IoU=0.7 & IoU=0.3 & IoU=0.5 & IoU=0.7 \\ \hline
			\multicolumn{7}{c}{C3D features} \\
			\hline
			MCN     & 39.35 & 21.36 & 6.43 & 68.12 & 53.23 & 29.70 \\
			TGN     & 47.43 & 29.01 & 10.34 & 75.32 & 59.17 & 37.54 \\
			ACRN    & 49.70 & 31.67 & 11.25 & 76.50 & 60.34 & 38.57 \\
			DEBUG   & 55.91 & 39.72 & - & - & - & - \\
			GDP     & 56.17 & 39.27 & - & - & - & - \\
			ABRL    & 55.67 & 36.79 & - & - & - & - \\
			TripNet & 48.42 & 32.19 & 13.93 & - & - & - \\
			TSP-PRL & 56.08 & 38.76 & - & - & - & - \\
			LGI     &  58.52 & 41.51 & 23.07   &   -   &   -   &   -    \\
			DRN     &  - & 45.45 & 24.36   &   -   &   \textbf{77.97} & 50.30    \\ 
			CTRL     &   -   & 14.00 &   -   &   -   &   -   &   -    \\
			QSPN   &   -   & 27.70 & 13.60 &   -   & 71.85 & 45.96  \\
			RWM-RL    &   -   & 36.90 &   -   &   -   &   -   &   -    \\
			SCDM     & 54.80 & 36.75 & 19.86 & 77.29 & 64.99 & 41.53  \\
			CBP     & 54.30 & 35.76 & 17.80 & 77.63 & 65.89 & 46.20  \\
			TSP-PRL  & 56.08 & 38.76 & - & - & - & - \\ 
			2D-TAN   & 59.45 & 44.51 & 26.54 & 85.53 & 77.13 & 61.96 \\ 
			CMIN     & - & 43.40 & 23.88 & - & 67.95 & 50.73 \\ 
			BPNet    & - & 42.07 & 24.69 & - & - & - \\
			RaNet    & - & \underline{45.59} & \underline{28.67} & - & 75.93 & \underline{62.97} \\
			FVMR     & \textbf{61.39} & 45.21 & 27.32 & \underline{85.98} &  77.10 & \textbf{63.44} \\ \hline
			Ours     & \underline{60.58} & \textbf{46.19} & \textbf{28.87} & \textbf{86.02} & \underline{77.86} & 60.28 \\ \hline
			\hline
			\multicolumn{7}{c}{I3D features} \\
			\hline
			ExCL     & \underline{62.30} & 42.70 & 24.10 & - & - & -  \\
			TMLGA    & 51.28 & 33.04 & 19.26 & - & - & -  - \\
			VSLNet   & \textbf{63.16} & \underline{43.22} & \underline{26.16} & - & - & -  \\ \hline
			\textbf{Ours} & 61.99 & \textbf{46.58} & \textbf{29.37} & \textbf{85.36} & \textbf{76.71} & \textbf{59.69} \\ \hline			
		\end{tabular}
	}\label{tab:acnet results}
	\vspace{-2mm}
\end{table}

\begin{table*}[t]
	\caption{Ablation Studies on TACoS and ActivityNet Captions}
	\begin{center}
		{	
			\begin{tabular}{l|ccc|ccc||ccc|ccc}
				\hline
				\multicolumn{1}{l|}{\multirow{3}{*}{Methods}} & \multicolumn{6}{c||}{{TACoS}} & \multicolumn{6}{c}{{ANetCap}}  \\
				\cline{2-13}
				& \multicolumn{3}{c|}{R@1} & \multicolumn{3}{c||}{R@5} & \multicolumn{3}{c|}{R@1} & \multicolumn{3}{c}{R@5} \\
				\cline{2-13}
				& IoU=0.1 & IoU=0.3 & IoU=0.5 & IoU=0.1 & IoU=0.3 & IoU=0.5 & IoU=0.3 & IoU=0.5 & IoU=0.7 & IoU=0.3 & IoU=0.5 & IoU=0.7 \\
				\hline
				backbone       & 50.42 & 39.25 & 26.55 & 77.47 & 60.15 & 46.33 & 58.60 & 41.93 & 24.02 & \underline{86.20} & \underline{77.98} & 54.45 \\
				Ours(w/o. v-c) & 51.82 & 40.55 & 28.88 & \textbf{76.95} & \underline{63.73} & 50.52 & 59.22 & 42.54 & 24.60 & \textbf{87.42} & \textbf{78.03} & 54.62 \\
				Ours(w/o. t-c) & \underline{53.45} & 41.40 & \underline{30.30} & 76.42 & 63.58 & 50.08 & \underline{60.89} & \underline{45.76} & \underline{27.89} & 85.85 & 77.48 & \underline{59.52} \\
				Ours(w/o. c-c) & 52.28 & \underline{42.38} & 30.18 & 77.53 & 62.25 & \underline{50.65} & 59.71 & 44.97 & 27.80 & 85.38 & 77.09 & 59.25 \\ \hline
				Ours(full) & \textbf{56.00} & \textbf{45.30} & \textbf{32.83} & \underline{76.60} & \textbf{64.38} & \textbf{52.68} & \textbf{60.58} & \textbf{46.19} & \textbf{28.87} & 86.02 & 77.86 & \textbf{60.28} \\ 			
				\hline
			\end{tabular}
		}\label{tab:ablationresults}
	\end{center}
\end{table*}

\subsection{Ablation Study} In this section, we take in-depth ablation studies to investigate the contribution of each main component in our proposed CCA method on TACoS and ActivityNet Captions datasets. Specifically, to perform complete ablation studies, we divide the consensus-aware interaction module into two attention-based modules: the visual-commonsense interaction module and the text-commonsense interaction module. We train CCA with the following configurations:
\begin{itemize}
	\item \textbf{backbone}: To prove the effect of each component, we directly map extracted visual and text features into common space to calculate similarity scores without any other modules.
\end{itemize}
\begin{itemize}
	\item \textbf{w/o. v-c}: We remove the visual-commonsense interaction module to verify whether commonsense knowledge can strengthen visual information.
\end{itemize}
\begin{itemize}
	\item \textbf{w/o. t-c}: We replace the text-commonsense interaction module with directly using encoded text features to further calculate matching scores.
\end{itemize}
\begin{itemize}
	\item \textbf{w/o, c-c}: To investigate the complementary role of original semantic information to the performance of CCA, only one common space is used for obtaining matching scores (only $n_i$ is used).
\end{itemize}
The ``full'' means the full CCA model. ~\tabref{ablationresults} summarizes the localization results in terms of ``R@\{1,5\}, IoU=\{0.1, 0.3, 0.5\}'' for TACoS and ``R@\{1,5\}, IoU=\{0.3, 0.5, 0.7\}'' for ActivityNet Captions. For the improvement of each branch, we have concrete analysis as follows:

{\noindent \bf{Effects of consensus-aware interaction module. }} We evaluate the effectiveness of the commonsense-aware interaction module by training our model only using no commonsense-guided visual or text features. From the results in~\tabref{ablationresults}, we can observe that, compared with backbone, no matter which type of modality features are fused with concept information can improve the performance of the model. Obviously, each interaction module has a positive effect on the VTG task. On TACoS, the full model outperforms ``w/o. t-c'' by gains of $ (2.55\%, 3.90\%, 2.53\%) $ on metrics ``R@1, IoU=\{0.1, 0.3, 0.5\}", and outperforms ``w/o. v-c'' by gains of $ (4.18\%, 4.75\%, 3.95\%) $ on the same metrics. For the ActivityNet Captions, the full model exceeds ``w/o. v-c'' by $ (1.36\%, 3.65\%, 4.27\%) $ in terms of ``R@1, IoU=\{0.3, 0.5, 0.7\}'' while achieves a significant $ 5.66\% $ absolute improvement in terms of ``R@5, IoU=0.7''. Besides, the full model also outperforms ``w/o. t-c'' by a large margin on all metrics. Obviously, the interaction between visual and commonsense knowledge leads to a improvement in performance, which proves that the exploitation of commonsense knowledge is significant to complementing visual information. Moreover, we can find that the text features are more effective when reinforced by commonsense knowledge. That is, commonsense knowledge can cooperate with visual and text features to obtain favorable results.

\begin{figure}[!t]
	\centering
	\includegraphics[width=1.0\linewidth]{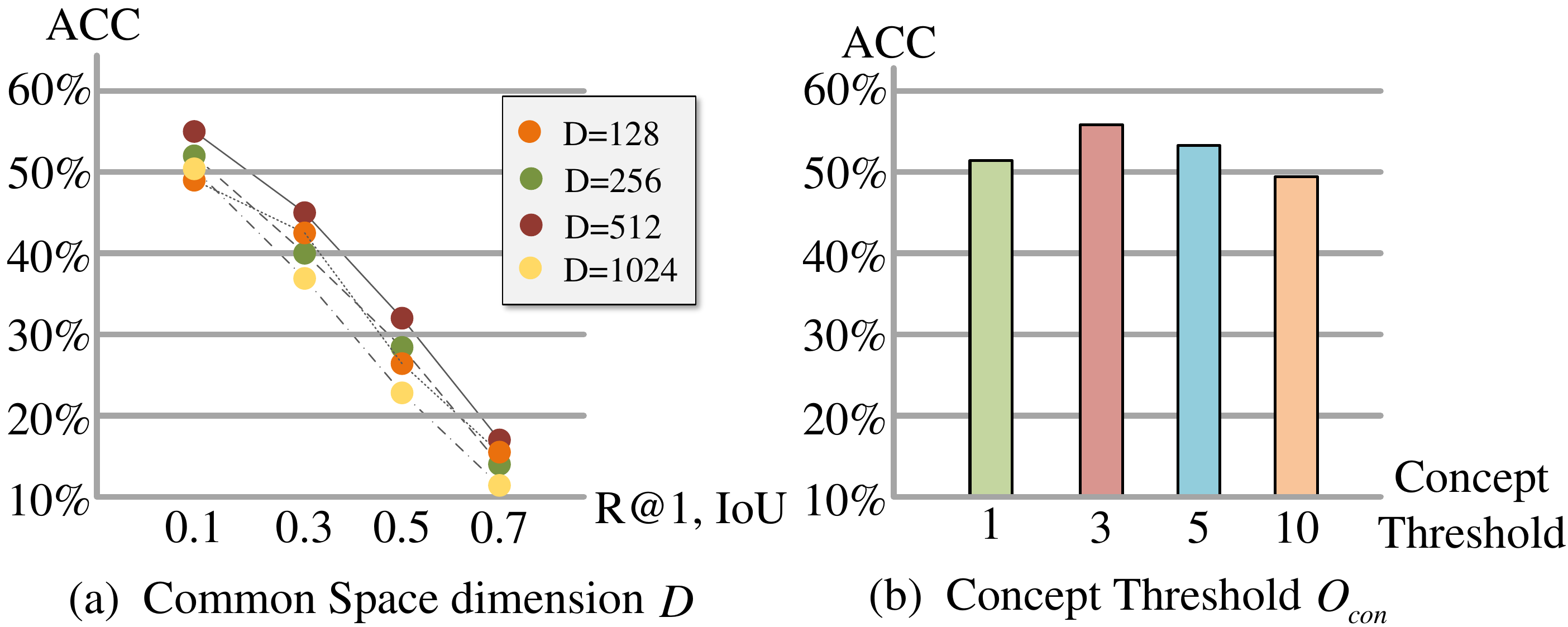}
	\caption{Performance comparison with different settings on common space dimension and concept selection on TACoS dataset.}
	\label{fig:ablation_result}
\end{figure}

{\noindent \bf{Effect of the complementary common space. }} To obtain discriminative representations, a complementary common space is designed to effectively compute matching scores while maintaining the original semantic information of the video and query. Here, a baseline ``w/o. c-c'' is designed to study the influence of the original information. From~\tabref{ablationresults}, the full model achieves a $ (3.72\%, 2.92\%, 2.65\%) $ improvement compared with ``w/o. c-c'' in terms of ``R@1, IoU=\{0.1, 0.3, 0.5\}'' on TACoS and achieves a $ (0.87\%, 1.22\%, 1.07\%) $ in terms of ``R@1, IoU=\{0.3, 0.5, 0.7\}'' on ActivityNet Captions. It proves that the original query can be a type of supplement for the consensus concept due to the fact that concepts are only words selected from the query.



{\noindent \bf{Effect of common space dimension. }} We investigate the influence of different dimensions of the learned common space on TACoS. As shown in \figref{ablation_result}(a), a too large dimension would lead to a higher cost of memory and times, which would also reduce the performance of our model. By contrast, a too small dimension always results in the lack of representation capability of learned common space. Therefore, a modest value of the common space dimension gets better performance.

{\noindent \bf{How many consensus concepts should be selected? }} In section III, we introduce the extraction of the commonsense concept by selecting the words with frequencies larger than $ O_{con} $. To explore the influence of the number of concepts, we validate the model performance with different $ O_{con} $ on TACoS.
As shown in \figref{ablation_result}(b), it is obvious that with the addition of the threshold value, the performance of our model does not always improve. As a result, we set the threshold to 3.
\begin{figure*}[!t]
	\centering
	\includegraphics[width=1.0\linewidth]{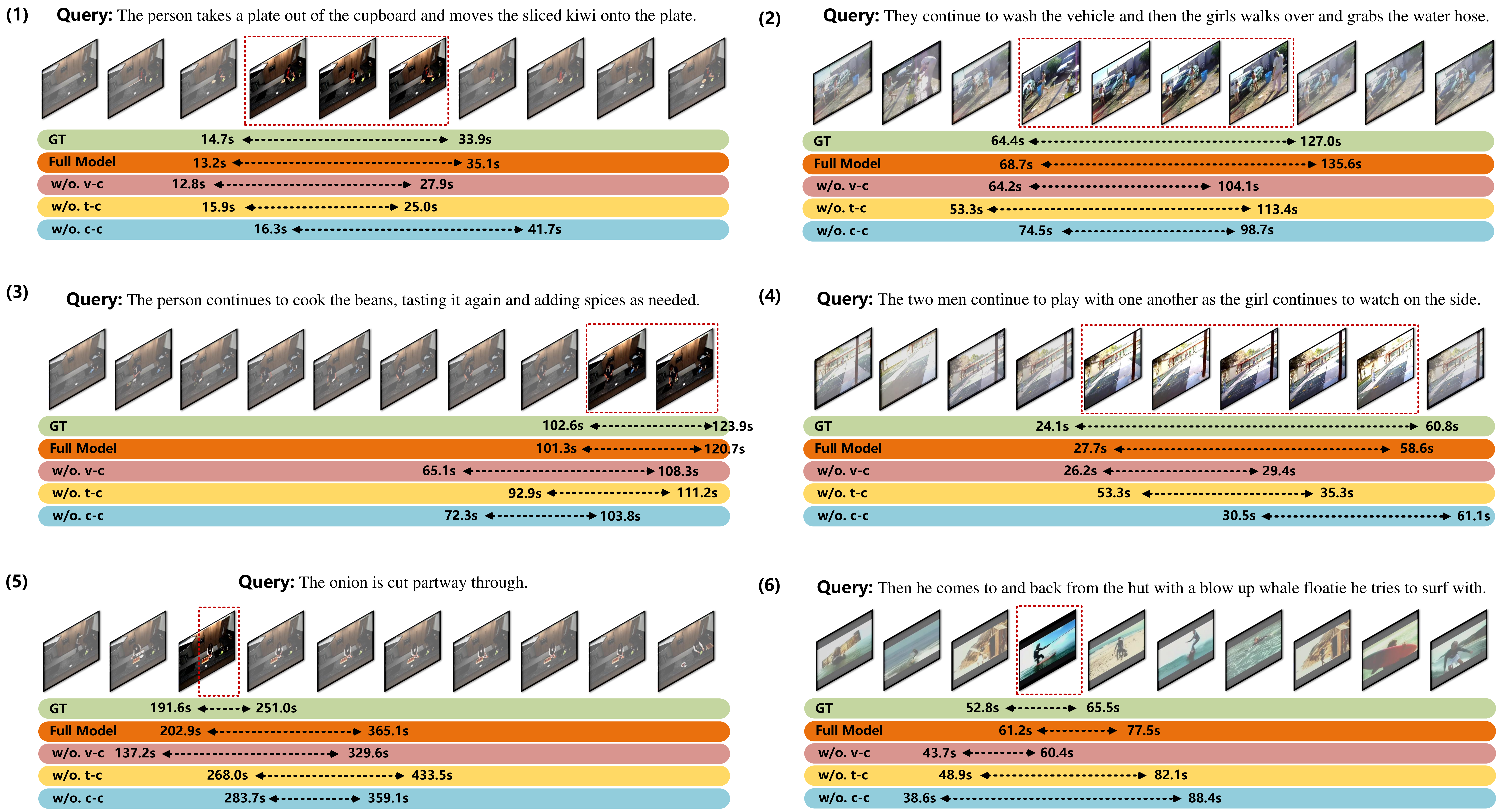}
	\caption{Visualization of the predictions of the CCA model and the ablation models on TACoS and ActivityNet Captions. \textbf{GT} is the ground truth moment, \textbf{Fuul Model} is the results of CCA. \textbf{w/o. v-c}, \textbf{w/o. t-c} and \textbf{w/o. c-c} are the results of three types of ablation models.}
	\label{fig:qualitative results}
\end{figure*}
\subsection{Qualitative Results}
To qualitatively validate the efficiency of our proposed CCA method, several typical examples are shown in \figref{qualitative results} for qualitative comparison. Besides the CCA method, three baselines, \ie, CCA (w/o. v-c), CCA (w/o. t-c), and CCA (w/o. c-c), are also validated for deep insight into the impact of different main components on localization performance. In \figref{qualitative results}, examples 1, 3, and 5 are from TACoS, while examples 2, 4, and 6 are from ActivityNet Captions. In examples 1-4, CCA (full) performs the results almost the same as ground truth, while the other three baselines output worse results than the full model. The performance in complex environments validates that our proposed CCA can accurately capture and comprehend the rich cross-modal relationship to a certain extent. Obviously, the ablation models predict the boundaries with a larger distance to ground truth compared with the full model. Since they either lack the crucial commonsense knowledge as a bridge between visual and text modalities or do not have a limit from original semantics. The result shows that the learning of commonsense knowledge plays an essential role in enhancing the discriminative ability of different modalities, and the original semantics are essential for better performance.

While our method can localize the correct temporal segment for most videos, failed samples also exist in some cases. As shown in \figref{qualitative results}, examples 5 and 6 present the cases of failing to localize the correct video moments. In example 5, CCA wants to localize the video segment that describes ``The onion is cut partway through''. As a whole, the query has too less semantic information for retrieval, which leads to an incorrect result. Moreover, ``partway'' is not a commonsense concept of TACoS. That is, CCA only pays more attention to the action ``cut'', and finally fails. In example 6, there is the same situation that the word ``whale'' and ``floaties'' are not commonsense concepts, which makes the CCA model not able to comprehend the complex and unacquainted video content. Note that there is a large distance between the prediction and the ground truth in sample 5, which also demonstrates that the quality of generated proposals also limits the performance of our proposed CCA.

%% file: conclus.tex
In this paper, we propose a novel Commonsense-aware Cross-modal Alignment (CCA) framework to achieve fast video temporal grounding, which incorporates commonsense-guided visual and text representations into a complementary common space and learns efficient commonsense-aware cross-modal alignment. We first utilize a multi-modal feature extractor to obtain different modality features. Specially, we select a set of representative words as commonsense concepts according to their occurrence frequencies. After that, a commonsense-aware interaction module is designed to obtain discriminative visual and text representations. Finally, we leverage a complementary common space to align visual and text representations to calculate their matching scores for temporal grounding. Experimental results of our proposed CCA method demonstrate that it achieves competitive performance when compared with other state-of-the-art methods on two benchmark datasets.

Three perspectives will be considered for future work. First, considering the common disadvantage of proposal-based methods, it is necessary to regress the boundaries of the predicted results, which could improve the accuracy of the prediction. Second, We believe that it is very challenging to localize such complex moments without richer commonsense knowledge. To improve the efficiency of knowledge utilization and the generalization of our model, knowledge graph, external commonsense database, and the pre-training process could also be taken into consideration. Finally, more sufficient and appropriate interactions between commonsense knowledge and visual or text features need to be explored such as modulated attention mechanism and  casual reasoning.